# Frequency-Domain Latent Attention Gating for Cross-Domain Token Aggregation

Kewei Li[1,2], Rongying Zhang[3], Xueli Wang[1,2], Xiwen Gong[4], Zhongjian Wang[5], Lan Huang[1,2], Ruochi Zhang[1,2,#], Fengfeng Zhou[1,2,#].

1 College of Computer Science and Technology, Jilin University, Changchun, 130012, China.

2 Key Laboratory of Symbolic Computation and Knowledge Engineering of Ministry of Education, Jilin University, Changchun, 130012, China.

3 Institute for Quantitative and Computational Biology, University of California, Los Angeles, 90024, United States of America.

4 Greenwich High School , Greenwich, CT, 06830, United States of America.

5 BCPM Data Limited, Chengdu 610041, China.

# Correspondence may be addressed to Drs. Ruochi Zhang (zrc720@gmail.com), and Fengfeng Zhou (FengfengZhou@gmail.com or ffzhou@jlu.edu.cn).

## Abstract

Token aggregation is a common bottleneck in models that map token representations to sample-level predictions, yet most pooling methods operate only in the original token domain. We propose FLaG, a plug-in aggregation module that transforms token representations with the real FFT, summarizes spectral components with learnable latent queries, applies a channel-wise gate, and reconstructs enhanced time-domain tokens for final pooling. We evaluate FLaG on antimicrobial peptide (AMP) activity prediction with ESM2, image classification with ResNet18 on CIFAR-10 and CIFAR-100, and text classification with RoBERTa on IMDB and GLUE. FLaG achieves its clearest gains on the ESM2-8M antimicrobial peptide tasks and on CIFAR-100, while remaining competitive with strong text baselines on IMDB and GLUE. Then we probe its behavior on the AMP setting with band knockouts, gate summaries, residue perturbations, latent-query readouts, and structure-proxy stratification. We find that low-frequency bands contribute the most overall, and the remaining higher-band pattern is more sample-specific. The gate acts as a broadly shared spectral reweighting stage and the cross-attention patterns are sample-specific with mild query-wise differentiation, and higher-helix peptides exhibit stronger average spectral sensitivity in both bacteria. The supplementary materials, source code and data are released at https://www.healthinformaticslab.org/supp/ and https://github.com/Kewei2023/AMPCliff/tree/FLaG.

Keywords: Frequency-domain token aggregation; latent attention gating; spectral re-expression; cross-domain pooling; antimicrobial peptide prediction.

# 1. Introduction

Modern neural architectures for antimicrobial peptide analysis, computer vision, and natural language processing share a common representation bottleneck: they must compress a variable-length sequence of token representations into a single fixed-length vector before task-specific prediction. This final aggregation step, which we refer to as token pooling, is easy to overlook because it sits after the encoder, yet it can materially affect what information survives to the prediction head. For antimicrobial peptide (AMP) models such as ESM2 [1], pooling reduces per-residue embeddings to a peptide-level representation for activity prediction; for vision models, spatial tokens or feature maps are pooled before classification; and for language models such as RoBERTa [2], token sequences are aggregated into sentence-level embeddings for downstream tasks.

Existing pooling rules are usually defined only in the original token domain. Simple reducers such as mean pooling, max pooling, and last-token selection remain widely used [2], [3], while attention-based pooling [4] learns token importance weights without changing the representation basis before the final reduction step.

However, in antimicrobial peptide prediction, we observed that simple pooling strategies can substantially affect the token representations in the final encoder layer. Specifically, mean pooling tends to make token representations more homogeneous and may even induce representation collapse, whereas max pooling preserves a higher degree of representation diversity. Although attention pooling can also maintain a certain level of diversity among final-layer token representations, its overall predictive performance is inferior to that of max pooling. These observations lead to two central questions. *First, can we develop a pooling mechanism that is better suited than max pooling for antimicrobial peptide prediction? Second, can this mechanism generalize effectively across different domains?*

Recent protein-specific pooling work points in a similar direction. EvoPool [5] argues that pooling residue embeddings from a single sequence can leave useful evolutionary information underexploited, and addresses this by constructing a fixed-size evolutionary anchor from homologous sequences and using that anchor to guide aggregation through sliced Wasserstein geometry. This is an appealing solution for protein tasks with accessible homolog context, but it also highlights a practical limitation: such external anchors are not available in many settings, including non-protein domains and protein tasks where homologs or MSAs are sparse, unavailable, or undesirable at inference time. This motivated us to look for an internal, domain-agnostic alternative.

We therefore turn to the frequency domain as a learned coordinate system that does not rely on homologs. FLaG first re-expresses token representations in spectral coordinates, then applies latent attention and channel-wise gating before reconstructing enhanced time-domain tokens for final pooling. The pipeline has four stages: (1) transform token representations with the real FFT; (2) apply learnable latent queries with multi-head attention to summarize task-relevant spectral components; (3) generate a channel-wise gate from the latent summary to modulate frequency amplitudes; and (4) reconstruct enhanced time-domain tokens via the inverse FFT and apply standard pooling. *Our goal is to ask whether frequency-domain re-expression can produce a stronger and more stable aggregation design, especially on tasks where sequence-level structure should matter after encoding.*

We test that hypothesis along an explicit evidence chain. First, we evaluate FLaG on the antimicrobial peptide tasks that motivated the method and then on image and text benchmarks to measure how far the same aggregation bias transfers across domains, Second, we analyze the AMP setting with five antimicrobial peptide-side probes: sequence-frequency band knockout, gate spectral effect analysis, single-residue perturbation, latent-query readout, and structure-proxy stratification. These analyses show low-frequency-dominant contribution patterns with sample-specific residual higher-band effects, a broadly shared spectral gate whose post-gate energy remains most concentrated in the lowest-frequency bands, residue-response profiles that remain substantially more differentiated than under mean pooling, sample-specific cross-attention patterns with mild query-wise differentiation, and stronger average sensitivity in higher-helix peptide buckets.

Our contributions are:

- we propose FLaG, a frequency-domain latent attention and gating pooling framework that enhances token aggregation through spectral re-expression, latent-query summarization, channel-wise gating, and inverse reconstruction, without requiring homologs, MSAs, or external anchors.
- we validate FLaG across antimicrobial peptide prediction, computer vision, and natural language processing tasks, showing that the proposed pooling mechanism provides a transferable aggregation bias across domains.
- we conduct five peptide-side probes, demonstrating that FLaG improves prediction on links its gains to informative spectral and residue-level representation patterns.

# 2. Related Work

## 2.1 Token Pooling Methods

Simple aggregation methods—mean pooling, max pooling, and last-token selection [1]—remain the most widely used pooling strategies due to their simplicity and zero-parameter overhead. Attention pooling [2] learns token importance weights through a trainable scoring function in the original token domain. Latent attention extends this by using multiple learnable query vectors, enabling richer summarization [3]. Multi-Layer Trainable Pooling (MLTP) aggregates across hidden layers before token pooling [4]. Sliced Wasserstein pooling approximates the distribution of token embeddings via random projections and quantiles [6].

## 2.2 Frequency-Domain Methods in Deep Learning

Spectral analysis has been applied to neural networks in several contexts. Spectral analyses of neural networks have shown that models often fit low-frequency components earlier in training [7]. In graph neural networks, spectral convolution operates on the graph Laplacian's frequency components [8]. For time series, frequency-domain representations improve forecasting by capturing periodicity [9]. This combination of learned latent attention and channel-wise spectral gating defines a distinct design point for cross-domain token aggregation.

## 2.3 Protein Representation Learning

Antimicrobial peptide language models such as ESM2 [10] generate per-residue embeddings that must be pooled for sequence-level tasks. The choice of pooling can significantly affect activity prediction performance [5], [11], [12]. Recent protein-specific work has further suggested that pooling can benefit from an explicit reference structure rather than from single-sequence aggregation alone. EvoPool [5], for example, constructs a fixed-size evolutionary anchor from homologous sequences and uses that external anchor to guide protein-level aggregation. This is a compelling design for protein tasks with accessible homolog context, but it is not directly transferable to domains such as images and text, or even to antimicrobial peptide settings where a suitable homolog set is unavailable at inference time.

# 3. Method

## 3.1 Problem Formulation

Given a token sequence $\mathbf{X} \in \mathbb{R}^{T \times D}$, where $T$ is the sequence length and $D$ is the hidden dimension, sequence pooling produces a fixed-length vector $\mathbf{z} \in \mathbb{R}^{D}$. An optional binary attention mask $\mathbf{m} \in \{0,1\}^{T}$ indicates valid positions. The pooling operation should respect the validity mask, avoid dependence on padded positions, and compress the sequence while preserving task-relevant information.

## 3.2 Overview

FLaG consists of four stages: frequency transformation, latent attention, frequency gating, and time-domain pooling. Figure 1 illustrates the complete pipeline.

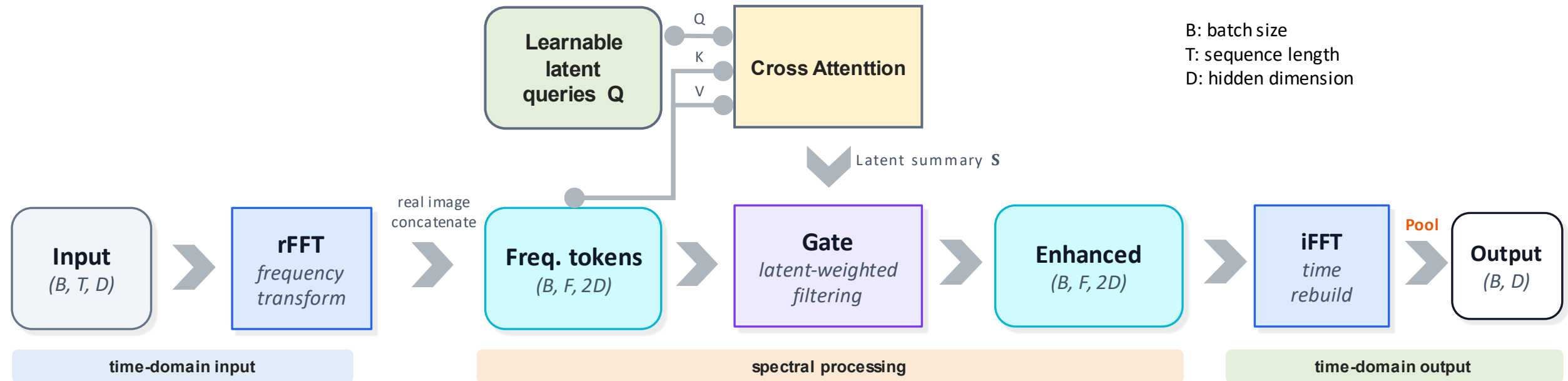

**Figure 1. Architecture of FLaG. Token representations are transformed with the real FFT and converted to frequency tokens by concatenating their real and imaginary parts. Latent queries operate in spectral space to produce the gate context, and the resulting channel-wise gate is shared across frequency bins before inverse reconstruction and final pooling.**

## 3.3 Design Intuition

### 3.3.1 From Token-Domain Pooling to Frequency-Domain Re-expression

Existing token pooling methods operate in the time domain [2], [10], [13]. Given an input sequence, attention-based pooling computes token importance weights based on pairwise dot-product similarity. However, in peptide and protein sequences, functionally relevant signals are often governed by local motifs, critical residues, and their context-dependent interactions [12]. Existing general-purpose pooling strategies usually compress residue-level representations into a single global vector, which can obscure fine-grained sequence information and thereby constrain performance in downstream functional prediction tasks [12]. FLaG is also informed by a second design intuition drawn from antimicrobial peptide-specific pooling work: pooling can improve when the model is given an explicit reference structure rather than being forced to aggregate directly in raw token coordinates. Where EvoPool [5] derives that reference from homolog-based evolutionary anchors, FLaG instead seeks a domain-agnostic internal reference frame by re-expressing the token sequence in spectral coordinates. This design begins with a basis change: it re-expresses the sequence in spectral coordinates before the final pooling stage.

The Real FFT transforms the signal into frequency components:

$$\widehat{\mathbf{X}}(f) = \sum_{t=0}^{T-1} \mathbf{x}(t) \cdot e^{-i\frac{2\pi}{T}ft}, \quad f \in \{0, \dots, F-1\} \tag{1.}$$

A basic property of the Fourier transform is that each frequency component depends on the full time-domain sequence [9]. This makes frequency space an alternative coordinate system for summarizing the full sequence before reduction. Whether that representation is useful is task-dependent and is therefore assessed empirically in Section 4.

### 3.3.2 Spectral Basis as a Design Choice

A well-known property of the Fourier transform is the shift theorem: if the input is shifted by $\tau$, the frequency representation becomes $\widehat{\boldsymbol{X}}(f)e^{-i\frac{2\pi f\tau}{T}}$[14]. While the phase rotates, the magnitude $|\widehat{\boldsymbol{X}}(f)|$ remains unchanged. This observation does not by itself imply an empirical advantage for FLaG, because the model operates on concatenated real and imaginary parts rather than on magnitudes alone. It does, however, motivate treating spectral coordinates as a viable intermediate representation rather than as a performance guarantee. FLaG leverages this design choice through three mechanisms. First, the real-imaginary concatenation preserves full spectral information while enabling processing by standard neural network layers. Second, the learnable latent queries in Eq. 4 provide a compact learned interface for attending to frequency tokens, rather than relying on hand-designed frequency bands [15]. Third, the gating mechanism performs channel-wise spectral modulation, allowing the model to preserve or amplify selected spectral components before reconstruction.

### 3.3.3 Latent Summarization and Gating as Learned Frequency Selection

The latent queries introduce an information bottleneck between the full frequency space (dimension $F \times 2D$) and the pooled output (dimension $D$), where $L \ll F$. The learned latent interface was partly inspired by anchor-style pooling ideas such as EvoPool [5], but the role is different here: instead of comparing a query sequence to homolog-derived external anchors, the latent queries summarize an internal frequency-domain coordinate system that is available even when no homolog set exists. One possible intuition is that the latent queries and gate together act as a learned frequency-selection layer: they summarize the spectrum, derive a task-conditioned modulation signal, and then reconstruct an updated token sequence for final pooling. The usefulness of that bias is ultimately an empirical question, and the clean-data results in Section 4.3.2 and Section 4.3.4 should be interpreted as evidence that this design can help on some domains.

## 3.4 Process

### 3.4.1 Frequency Transformation

We apply the Real Fast Fourier Transform (rFFT) along the sequence dimension to obtain frequency-domain representations:

$$\hat{\mathbf{X}} = \mathrm{rFFT}(\mathbf{X} \odot \mathbf{m}) \in \mathbb{C}^{F \times D} \tag{2.}$$

where $F = \lfloor \frac{T}{2} \rfloor + 1$ is the number of frequency bins and $\mathbf{m}$ is the optional attention mask (broadcast over $D$). In practice, variable-length batches are zero-padded to a uniform batch length, masked before transformation, and transformed at that padded length so that the FFT can be applied on a rectangular tensor. We represent the complex spectrum as concatenated real and imaginary parts:

$$\mathbf{F} = \left[\mathrm{Re}(\hat{\mathbf{X}}) \parallel \mathrm{Im}(\hat{\mathbf{X}})\right] \in \mathbb{R}^{F \times 2D} \tag{3.}$$

This representation preserves both magnitude and phase information across all frequency bins.

### 3.4.2 Latent Attention in Frequency Space

We introduce $L$ learnable latent queries $\mathbf{Q} \in \mathbb{R}^{L \times 2D}$ that attend to the frequency tokens $\mathbf{F}$ via multi-head attention:

$$\mathbf{A} = \mathrm{MHA}(\mathbf{Q}, \mathbf{F}, \mathbf{F}) \in \mathbb{R}^{L \times 2D} \tag{4.}$$

The latent queries serve as compact learned "frequency detectors" that extract task-relevant spectral patterns. We apply pre-norm residual connections and a feed-forward network:

$$\mathbf{H} = \mathrm{LayerNorm}\big(\mathbf{Q} + \mathrm{Dropout}(\mathbf{A})\big) \tag{5.}$$

$$\mathbf{S} = \mathrm{LayerNorm}\Big(\mathbf{H} + \mathrm{Dropout}\big(\mathrm{FFN}(\mathbf{H})\big)\Big) \tag{6.}$$

where FFN consists of two linear layers with GELU activation and dropout.

### 3.4.3 Frequency Gating

The latent summary $\mathbf{S}$ is averaged across latent slots to produce a frequency-aware gate:

$$\bar{\mathbf{s}} = \frac{1}{L} \sum_{l=1}^{L} \mathbf{S}_l \in \mathbb{R}^{2D} \tag{7.}$$

$$\mathbf{g} = \sigma(\mathrm{MLP}(\bar{\mathbf{s}})) \in [0,1]^{2D} \tag{8.}$$

where $\sigma$ is the sigmoid function and MLP is a two-layer network with GELU. The gate modulates frequency tokens with a residual connection:

$$\tilde{\mathbf{F}} = \mathbf{F} \odot (1 + \mathbf{g}) \tag{9.}$$

The residual formulation ( $1 + \mathbf{g}$ rather than $\mathbf{g}$ alone) ensures that the gate can amplify important frequency components ($g > 0$) while preserving the original signal when the gate is inactive ($g \approx 0$).

### 3.4.4 Time-Domain Reconstruction and Pooling

The modulated frequency tokens are split into real and imaginary parts,

$$\tilde{\mathbf{F}} = \left[\tilde{\mathbf{F}}^{(R)} \parallel \tilde{\mathbf{F}}^{(I)}\right] \tag{10.}$$

converted back to complex form, and transformed via inverse FFT:

$$\tilde{\mathbf{X}} = \mathrm{irFFT}\left(\tilde{\mathbf{F}}^{(R)} + i\tilde{\mathbf{F}}^{(I)}\right) \in \mathbb{R}^{T \times D} \tag{11.}$$

Standard pooling (such as max pooling) is then applied to the reconstructed time-domain tokens:

$$\mathbf{z} = \mathrm{Pool}(\tilde{\mathbf{X}}, \mathbf{m}) \tag{12.}$$

We support max pooling, mean pooling, and attention pooling as the final aggregation. A final linear projection produces the output:

$$\mathbf{z}_{\mathrm{out}} = \mathbf{W}\,\mathrm{LayerNorm}(\mathrm{Dropout}(\mathbf{z})) \tag{13.}$$

## 3.5 Complexity Analysis

Let $T$ be the sequence length, $D$ the hidden dimension, $L$ the number of latent queries, and $H$ the number of attention heads. The rFFT/iFFT each contribute $O(TD\log T)$. Latent attention over frequency tokens contributes $O(FLD)$, ignoring constant factors from the multi-head projections and feed-forward layers, where $F \approx T/2$. The dominant arithmetic terms are therefore $O(TD\log T + TLD)$ when $L \ll T$. Compared with lightweight token-domain pooling rules, FLaG adds an FFT-dependent cost whose practical impact depends on sequence length and implementation.

# 4. Experiments

## 4.1 Experimental Setup

### 4.1.1 Domains and Datasets

We evaluate FLaG across three primary domains and then report auxiliary diagnostic analyses on antimicrobial peptide-side mechanism probes:

- Antimicrobial peptide Activity Prediction. Using ESM2 (8M and 35M parameters) [10] as the backbone, we evaluate antimicrobial peptide activity prediction on two species, *E. coli* and *S. aureus*. We follow AMPCliff [16] in both dataset construction and evaluation, including its activity-cliff-aware (AC) split, which is designed to probe generalization under activity-cliff constraints rather than under a purely random partition. We report both Recall@50 and RMSE. Recall@50 is motivated by the practical preference for top-candidate retrieval in peptide screening and, as defined in AMPCliff [16], measures the overlap between the 50 peptides with the best predicted MIC values and the 50 peptides with the best ground-truth MIC values. RMSE is reported as a complementary metric because it directly reflects the regression loss optimized during training and captures absolute prediction quality.

- Image Classification. We follow the evaluation protocol used in the universal pooling work [17]. In particular, ResNet18 [18] is used as the backbone, and its final spatial feature map is reshaped into a sequence of tokens so that pooling can be studied as a sequence aggregation problem rather than only as a fixed spatial reduction step. We evaluate on CIFAR-10 and CIFAR-100 [19], and report top-1 accuracy in percentage as the main metric.

- Language Classification. Using RoBERTa-base [13] as the backbone, we evaluate text-side pooling on binary sentiment classification with IMDB [20] and on the MNLI, QNLI, and SST-2 classification tasks from GLUE [21]. Performance is reported as accuracy in percentage for all text benchmarks.

These auxiliary analyses are descriptive rather than new primary benchmarks, and they are included to the behavior of the learned frequency-domain module on the AMP setting.

### 4.1.2 Baselines

We compare against seven primary baselines that cover the main pooling families used in our experiments, including mean pooling, max pooling, last-token selection (last) [1], attention pooling (attn) [2], latent attention (latent_attn) [3], sliced Wasserstein pooling (swe_ot) [6], and Multi-Layer Trainable Pooling (mltp) [4].

### 4.1.3 Implementation Details

All main experiments are repeated over 10 random seeds. For FLaG, we set the number of latent queries to 8, the number of attention heads to 4, the dropout rate to 0.1, enable the gate residual connection, and use max pooling as the final time-domain aggregation by default. Within each domain, we reuse the baseline training recipe (optimizer family and learning-rate schedule) so that the pooling comparison does not depend on retuning unrelated optimization components. No domain-specific hyperparameter tuning is performed for FLaG. Pooling is inserted immediately before the task-specific prediction head in all three domains: over final-layer residue embeddings for ESM2, over the flattened final spatial feature map for ResNet18, and over the final hidden-state token sequence for RoBERTa. For the text tasks, the main IMDB and GLUE tables compare aggregation rules over the same final hidden-state sequence so that the effect of the reducer can be isolated from unrelated classifier changes. For IMDB, we additionally report three RoBERTa-native sentence-level references: the first-token hidden state, the built-in RoBERTa pooler, and the default RoBERTa sequence-classification head.

## 4.2 Main Results

### 4.2.1 Antimicrobial peptide Activity Prediction

Table 1 presents results for antimicrobial peptide activity prediction using ESM2.

**Table 1. Antimicrobial peptide activity prediction results with ESM2. All results are averaged over 10 seeds. Bold and underline in the RMSE column indicate the best and second-best values, respectively. Recall@50 is reported as a secondary summary metric, while RMSE is the primary metric. Lower is better for RMSE.**

| | | | RMSE ↓ | | Recall@50 ↑ |
|---|---|---|---|---|---|
| Model | Dataset | Pooling | Mean | Std | Mean |
| ESM2-8M | *E. coli* | mean | 0.578 | 0.007 | <u>17.7</u> |
| | | max | 0.576 | 0.026 | 17.0 |
| | | attn | 0.583 | 0.015 | 16.7 |
| | | last | 0.573 | 0.012 | 16.8 |
| | | swe_ot | <u>0.568</u> | 0.009 | 17.3 |
| | | mltp | 0.601 | 0.023 | 15.7 |
| | | latent_attn | 0.577 | 0.016 | 16.7 |

| | | | RMSE ↓ | | Recall@50 ↑ |
|---|---|---|---|---|---|
| | | FLaG | **0.562** | 0.010 | **18.6** |
| ESM2-8M | *S. aureus* | mean | 0.577 | 0.019 | 16.2 |
| | | max | <u>0.552</u> | 0.015 | **18.4** |
| | | attn | 0.576 | 0.019 | 16.3 |
| | | last | 0.563 | 0.006 | 17.8 |
| | | swe_ot | 0.567 | 0.014 | 16.4 |
| | | mltp | 0.592 | 0.017 | 15.3 |
| | | latent_attn | 0.570 | 0.013 | 15.5 |
| | | FLaG | **0.545** | 0.005 | <u>18.0</u> |
| ESM2-35M | *E. coli* | mean | 0.555 | 0.005 | 19.3 |
| | | max | <u>0.546</u> | 0.008 | <u>20.3</u> |
| | | attn | 0.555 | 0.017 | **20.9** |
| | | last | 0.554 | 0.010 | 20.3 |
| | | swe_ot | 0.556 | 0.011 | 21.0 |
| | | mltp | 0.561 | 0.012 | 18.1 |
| | | latent_attn | 0.560 | 0.014 | 18.9 |
| | | FLaG | **0.546** | 0.013 | **20.9** |
| ESM2-35M | *S. aureus* | mean | 0.559 | 0.009 | 17.7 |
| | | max | **0.548** | 0.006 | <u>18.3</u> |
| | | attn | 0.552 | 0.012 | 17.5 |
| | | last | <u>0.550</u> | 0.011 | 17.2 |
| | | swe_ot | 0.551 | 0.012 | 18.0 |
| | | mltp | 0.558 | 0.013 | 17.8 |
| | | latent_attn | 0.557 | 0.006 | 17.3 |
| | | FLaG | 0.554 | 0.013 | **18.6** |

On the 8M backbone, FLaG achieves the lowest mean RMSE on both peptide tasks, reaching 0.562 on *E. coli* and 0.545 on *S. aureus*. The mean RMSE is lower than mean pooling in both cases (0.578 to 0.562 and 0.577 to 0.545), suggesting that spectral pooling is particularly helpful when the backbone has limited excess capacity. The Recall@50 results are broadly consistent with this picture on *E. coli*, where FLaG also ranks first at 18.6, but they are slightly more mixed on *S. aureus*, where FLaG reaches 18.0 and remains close to the best-performing max pooling baseline at 18.4. Taken together, the 8M results indicate that FLaG offers the clearest overall advantage when the model capacity is relatively limited, with the strongest gains appearing in RMSE and a generally competitive ranking signal in Recall@50.

With ESM2-35M, the picture becomes more mixed. FLaG remains best on *E. coli*, but only by a margin of less than 0.001 RMSE over max pooling, and it no longer leads on *S. aureus*, where max pooling achieves the lowest RMSE (0.548) while FLaG remains competitive at 0.554, but FLaG attains the highest Recall@50 in this setting at 18.6. This larger-model regime therefore narrows the claim: the clearest protein-side gains appear in the lower-capacity setting, while with the bigger backbone the trade-off becomes more metric-dependent and leaves less room for any one pooling rule to dominate uniformly.

### 4.2.2 Image Classification

Table 2 presents results for image classification with ResNet18.

**Table 2. Image classification accuracy (%) with ResNet18 on CIFAR-10/100. Averaged over 10 seeds. Bold and underline in the accuracy columns indicate the best and second-best values, respectively.**

| | CIFAR-10 | | CIFAR-100 | |
|---|---|---|---|---|
| Pooling | Acc. | Std | Acc. | Std |
| mean | 95.94 | 0.100 | <u>76.78</u> | 0.307 |
| max | 95.77 | 0.189 | 76.61 | 0.360 |
| attn | 95.91 | 0.056 | 76.65 | 0.225 |
| last | 95.89 | 0.115 | 75.14 | 0.425 |
| swe_ot | 95.92 | 0.115 | 76.65 | 0.316 |
| mltp | 95.90 | 0.141 | 75.85 | 0.324 |
| latent_attn | <u>95.96</u> | 0.123 | 75.71 | 0.365 |
| FLaG | **96.01** | 0.167 | **77.20** | 0.139 |

FLaG achieves the highest mean accuracy on both CIFAR-10 (96.01%) and CIFAR-100 (77.20%). On CIFAR-100, FLaG improves over the strongest baseline mean (76.78%) by 0.42 percentage points, and over attention pooling (76.65%) by 0.55 points. FLaG also exhibits the lowest variance on CIFAR-100 (std = 0.139), roughly half that of mean and max pooling, indicating more stable predictions across different random initializations in these runs. Even against the strong FFT+gate simplification introduced later in the ablation study (see Table 6), the full model remains slightly better on CIFAR-100 (77.20% vs. 77.09%), indicating that latent summarization and the final pooling stage still matter once the frequency block itself is fixed.

On CIFAR-10, where all methods achieve high accuracy (>95.7%), the margins are smaller but FLaG still leads, demonstrating that frequency-domain pooling provides benefits even when the task is relatively easy.

### 4.2.3 Language Benchmarks

Table 3 presents the IMDB aggregation comparison for RoBERTa-base.

**Table 3. IMDB sentiment classification with RoBERTa-base. Averaged over 10 seeds. Bold and underline in the accuracy column indicate the best and second-best values, respectively.**

| Pooling | Acc. | Std |
|---|---|---|
| mean | **94.08** | 0.206 |
| max | 93.96 | 0.239 |
| attn | 93.77 | 0.360 |
| last | <u>94.04</u> | 0.260 |
| swe_ot | 94.00 | 0.177 |
| mltp | 93.95 | 0.184 |
| latent_attn | 93.95 | 0.176 |
| FLaG | **94.08** | 0.191 |

FLaG matches the best accuracy (94.08%) achieved by mean pooling in the interchangeable token-pooling comparison, while exhibiting lower variance (std = 0.191 vs. 0.206). It also remains slightly above the simplified FFT+gate control in headline accuracy (94.08% vs. 94.07%), even though the simplified variant has the lowest variance among the four main comparison points. This reinforces the broader text-side story: under the current evaluation protocol, frequency-domain processing behaves as a high-floor alternative rather than as a large-margin replacement for strong semantic baselines.

**Table 4. IMDB RoBERTa-native reference baselines. The first three rows use sentence representations native to the backbone: the final hidden state at the initial <s> token (first_token), the RoBERTa affine+tanh pooler applied to that token (roberta_pooler), and the default RoBERTa sequence-classification head**

**(roberta_cls_head). Averaged over 10 seeds. Bold and underline in the accuracy column indicate the best and second-best values, respectively.**

| Representation | Acc. | Std |
|---|---|---|
| first_token | 94.02 | 0.260 |
| roberta_pooler | **94.08** | 0.162 |
| roberta_cls_head | <u>94.06</u> | 0.074 |
| FLaG | **94.08** | 0.191 |

Table 4 adds the most natural RoBERTa-native references on IMDB. Against those sentence-level baselines, FLaG remains competitive rather than clearly superior: it is effectively tied with the built-in pooler, slightly above the default classification head in headline accuracy, and slightly above the raw first-token representation. Those gaps are small enough that we do not interpret them as a distinct text-side win. Instead, the main value of this comparison is completeness: once the native RoBERTa references are included, the IMDB evidence still supports the same conservative conclusion that FLaG is a viable high-floor alternative on text.

**Table 5. GLUE validation accuracy (%) with RoBERTa-base over 10 seeds. Bold indicates the best result in each column.**

| Pooling | MNLI | QNLI | SST-2 |
|---|---|---|---|
| mean | **87.53** | **92.65** | 91.02 |
| attn | 87.37 | 92.64 | **91.11** |
| latent_attn | 87.34 | 92.48 | 90.95 |
| FLaG | 87.20 | 92.54 | 90.85 |

Table 5 reports the corresponding token-pooling comparison on GLUE and supports a more nuanced claim than "wins everywhere". Unlike IMDB, this GLUE table is intentionally limited to reducers applied to the shared final hidden-state token sequence; RoBERTa-native sentence-level references are not part of the present GLUE comparison. On clean text classification, mean pooling remains strongest on MNLI and QNLI, while attention pooling is best on SST-2; FLaG stays within 0.32 points of the best method on all three tasks.

### 4.2.4 Relative to Latent Attention

FLaG generally improves upon latent_attn, its closest time-domain counterpart, on the antimicrobial peptide and image benchmarks while remaining close on the text benchmarks (Tables 1, 2, and 5). On CIFAR-100, the gap is 1.48 points (77.20% vs. 75.71%; Table 2). On ESM2-8M of *E. coli*, FLaG achieves 0.562 vs. 0.577 RMSE for latent_attn (Table 1). On GLUE, the two methods are close, with FLaG ahead on QNLI but slightly behind latent attention on MNLI and SST-2 (Table 5). On IMDB, FLaG also stays slightly above latent_attn in clean accuracy (94.08% vs. 93.95%; Table 3). Taken together, these comparisons suggest that the frequency transform and subsequent gating contribute behavior beyond simply adding another set of learned queries.

Since our original motivation comes from the peptide task, and FLaG shows its strongest performance there, we carry out the subsequent mechanism analysis on the peptide benchmarks.

## 4.3 Antimicrobial Peptide-Side Mechanism Probes

To clarify what the module learns on the AMP setting, we conducted five antimicrobial peptide-side probes on ESM2-8M for both *E. coli* and *S. aureus*: sequence-frequency band knockout, gate spectral effect, single-residue perturbation, latent-query readout, and structure-proxy stratification.

For the probes with a per-sample view, the main text reports results aggregated over 10 aligned seeds on the same 10 representative peptides, with five peptides from each bacterium. Since extending the full per-sample analysis to all peptides would be computationally expensive, we instead randomly selected 30 peptides from each of *E. coli* and *S.*

*aureus* for the mechanism probes; the complete galleries for these sampled peptides are provided in Supplementary Section S3.

### 4.3.1 Sequence-Frequency Band Knockout

For this probe, we follow a prior DCT-based spectral knockout protocol [22] rather than reusing FLaG's internal rFFT computation directly. Specifically, the hidden states at a chosen layer are transformed with the discrete cosine transform (DCT), one band is removed by a notch-style mask, and the perturbed representation is reconstructed by the inverse DCT. Accordingly, FLaG itself still operates with rFFT in the model, whereas DCT is used here only to implement a stable, literature-aligned band-knockout analysis on real-valued hidden states.

We denote the hidden representation at layer $\ell$ as $\mathbf{H}_i^{\ell} \in \mathbb{R}^{T_i \times D}$, where $T_i$ is sequence length of peptide $i$ and $D$ is hidden dimension. For each layer-band pair, we transform $\mathbf{H}_i^{\ell}$ along the sequence axis into frequency coefficients $\mathbf{Z}_i^{(\ell)}$ by DCT, remove one target band by a binary mask, and reconstruct the perturbed representation by the inverse transform.

Let $\boldsymbol{\mathcal{B}}_b$ be the index set of frequency bins in band $b$ $(b = 0, \dots, K-1)$. With mask $M_b$ that equals 0 on $\boldsymbol{\mathcal{B}}_b$ and 1 elsewhere, the perturbed coefficients and reconstruction are defined as:

$$\widetilde{\mathbf{Z}}_i^{\ell,b} = \mathbf{Z}_i^{\ell} \odot M_b, \qquad \widetilde{\mathbf{H}}_i^{\ell,b} = i\mathrm{DCT}\left(\widetilde{\mathbf{Z}}_i^{\ell,b}\right) \tag{14.}$$

To avoid confounding the knockout effect with trivial scale shrinkage, we apply energy matching with a small numerical stabilizer $\varepsilon > 0$:

$$\widetilde{\mathbf{H}}_i^{\ell,b} \leftarrow \widetilde{\mathbf{H}}_i^{\ell,b} * \left( \frac{\left\|\mathbf{H}_i^{\ell}\right\|_F}{\left\|\widetilde{\mathbf{H}}_i^{\ell,b}\right\|_F + \varepsilon} \right) \tag{15.}$$

If $\hat{y}_i$ is the baseline prediction and $\hat{y}_{i,\ell,b}^{\mathrm{ko}}$ is the prediction after knocking out band $b$ at layer $\ell$, we record the per-peptide response as

$$\Delta\mathrm{MSE}_{i,\ell,b} = \left(\hat{y}_{i,\ell,b}^{\mathrm{ko}} - \hat{y}_i\right)^2 - (\hat{y}_i - y_i)^2 \tag{16.}$$

The heatmaps report $\Delta\mathrm{MSE}_{i,\ell,b}$ after averaging over the 10 aligned seeds across both datasets.

As is shown in Figure 2, the band-knockout response aggregated over 10 aligned seeds is dominated by $b = 0$ (the lowest sequence-frequency band), and the largest responses recur in the middle-to-late layers, peaking around layer 3. At the same time, the remaining bands are not uniformly irrelevant: their residual contribution pattern varies across peptides, so the non-low-frequency response is better understood as sample-specific than as a single shared secondary profile. Some *E. coli* peptides retain a visible high-band tail response, but the overall picture is still one of low-frequency dominance rather than evenly distributed band usage.

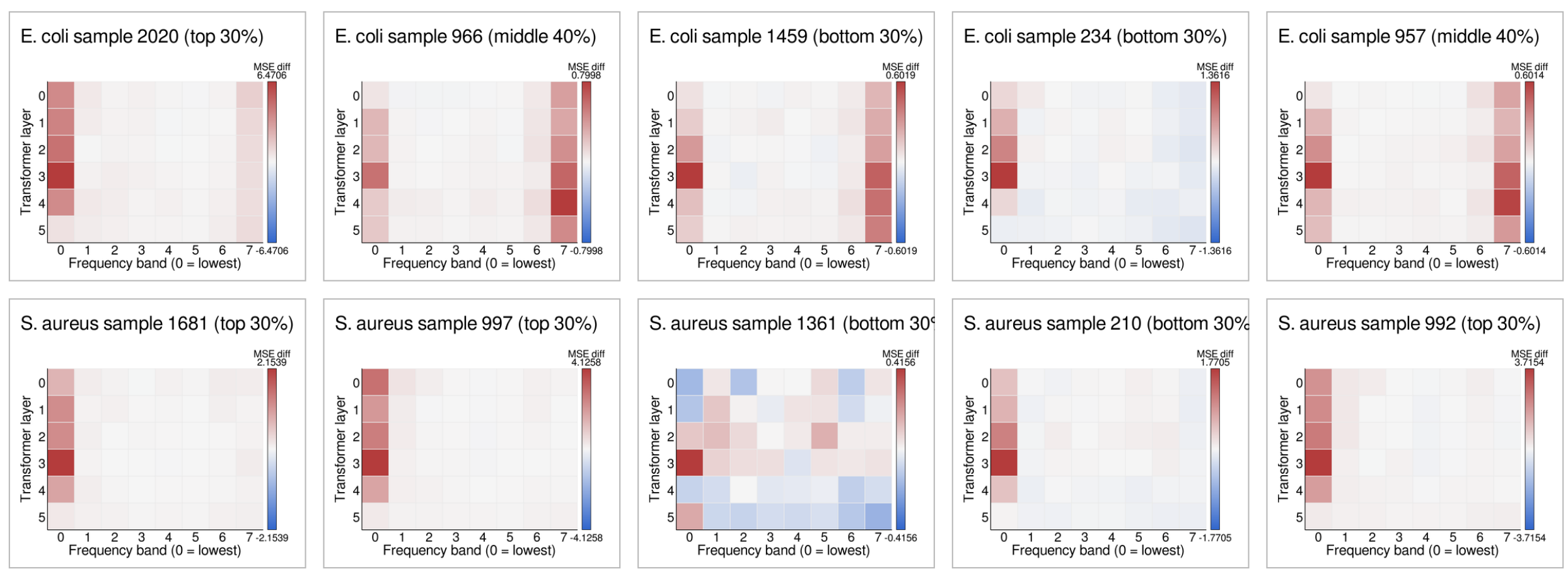

**Figure 2. Representative per-sample sequence-frequency band-knockout heatmaps on the ESM2-8M AMP tasks, aggregated over 10 aligned seeds. The 10 peptides were selected from the currently available amino-acid knockout subset so that the same cases can be followed across the antimicrobial peptide-side probes, with five peptides from each bacterium. In the 2 x 5 layout, the top row corresponds to *E. coli* and the bottom row corresponds to *S. aureus*. Within each panel, the horizontal axis shows the sequence-frequency band, ordered from lower to higher frequency, and the vertical axis shows the transformer layer. The text above each panel identifies the bacterium-specific AMP task and the sample identifier; when a parenthesized bucket label is present, it denotes the helix-propensity bucket assigned in Section 4.3.5 after ranking the 30-peptide manifest subset for that bacterium and partitioning it into the lowest 30%, middle 40%, and highest 30% of helix-propensity scores. Across both bacteria, the strongest perturbation response remains concentrated on the lowest sequence-frequency band and is typically largest in the middle-to-late layers.**

### 4.3.2 Gate Spectral Effect

The knockout probe identifies which bands matter to prediction, but it does not directly show how the learned gate reshapes the spectrum during standard inference. Here $\mathbf{F_i}$ the pre-gate frequency-token matrix for peptide $i$ defined in Eq.3, let $\mathbf{g_i}$ denotes the gate defined in Eq.8, and $\widetilde{\mathbf{F}}_\mathbf{i}$ denotes the gated frequency-token matrix defined in Eq.9. Writing $\mathbf{F}_{i,f}$ and $\widetilde{\mathbf{F}}_{i,f}$ for the tokens at frequency bin $f$ and $\boldsymbol{\mathcal{B}}_b$ for the set of bins assigned to band $b$, the implementation first computes the per-bin energies

$$e_{i,f}^{pre} = \left|\left|\mathbf{F}_{i,f}\right|\right|_2^2, \qquad e_{i,f}^{post} = \left|\left|\widetilde{\mathbf{F}}_{i,f}\right|\right|_2^2 \tag{17.}$$

and then stores the band-wise means

$$E_{i,b}^{pre} = \frac{1}{|\boldsymbol{\mathcal{B}}_b|} \sum_{f\,\in\,\boldsymbol{\mathcal{B}}_b} e_{i,f}^{pre}, \qquad E_{i,b}^{post} = \frac{1}{|\boldsymbol{\mathcal{B}}_b|} \sum_{f\,\in\,\boldsymbol{\mathcal{B}}_b} e_{i,f}^{post} \tag{18.}$$

The Figure 3 therefore show, for each peptide, the blue pre-gate bars $E_{i,b}^{pre}$ and the orange post-gate bars $E_{i,b}^{post}$ across the eight sequence-frequency bands. In the aggregated analysis across 10 seeds, the post-gate band energies $E_{i,b}^{post}$ are consistently higher than the corresponding pre-gate energies $E_{i,b}^{pre}$ across all eight sequence-frequency bands on both datasets. At the same time, the lowest-frequency bands remain the most energetic before and after gating. This pattern indicates that the gate broadly amplifies the spectrum while preserving the original low-frequency dominance, rather than redistributing energy toward a narrowly selected band.

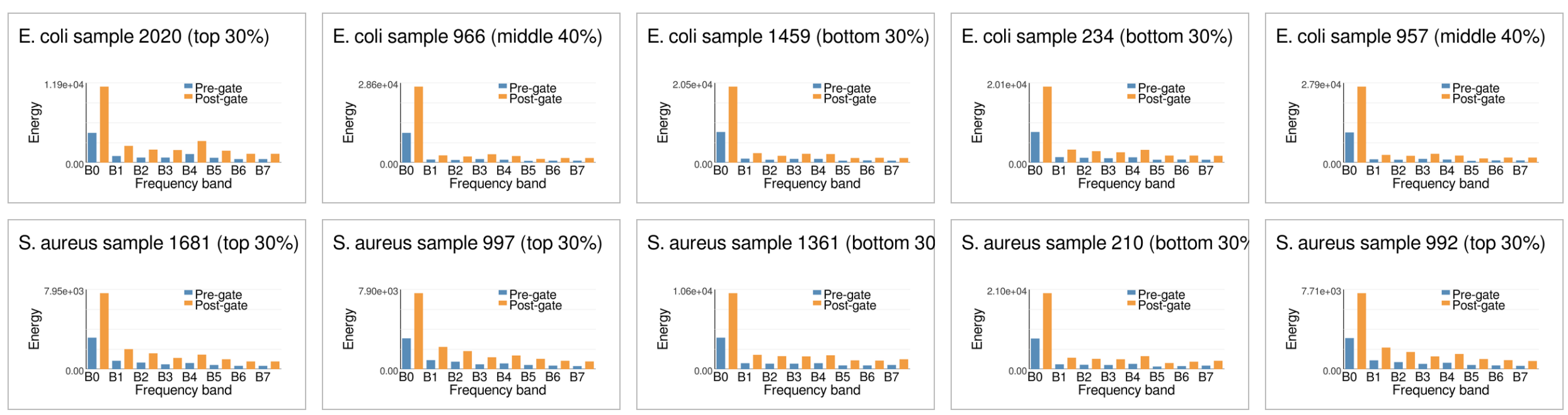

**Figure 3. Representative per-sample gate spectral summaries on the ESM2-8M AMP tasks, aggregated over 10 aligned seeds. The 10 peptides are arranged in the same 2 x 5 layout as in Figure 2, with *E. coli* on the top row and *S. aureus* on the bottom row. Within each panel, the horizontal axis shows the sequence-frequency band and the vertical axis shows band energy. The blue bars show the pre-gate band energy $E_{i,b}^{pre}$, and the orange bars show the post-gate band energy $E_{i,b}^{post}$ after the residual modulation step in Eq.9. The text above each panel identifies the bacterium-specific AMP task and the sample identifier; parenthesized bucket labels use the helix-propensity stratification summarized in Section 4.3.5, obtained by ranking the 30-peptide manifest subset within each bacterium and splitting it into the bottom 30%, middle 40%, and top 30% of**

**helix-propensity values. The local cases confirm the dataset-level result: FLaG broadly amplifies the spectrum, but does not produce a narrow band-specific gate.**

### 4.3.3 Single-Residue Knockout

This probe is implemented as a single-residue hidden-state knockout diagnostic rather than a generic perturbation analysis. For each peptide, we take the final encoder layer, mask one residue position at a time, rerun the model, and measure the absolute prediction change relative to the unmodified forward pass. The resulting response profile shows how strongly localized residue-level information survives the final pooling step. The results are shown in Figure 4.

Across representative peptides in both bacteria, after aggregating the single-residue knockout response over 10 aligned seeds, mean pooling remains comparatively flat, whereas max pooling, attention pooling, and FLaG preserve more differentiated position-wise response profiles. This indicates that aggregation choice materially affects how much localized residue-level variation is retained in the final sequence-level prediction, with mean pooling as the flattest reference and FLaG usually showing a substantially larger response range across positions.

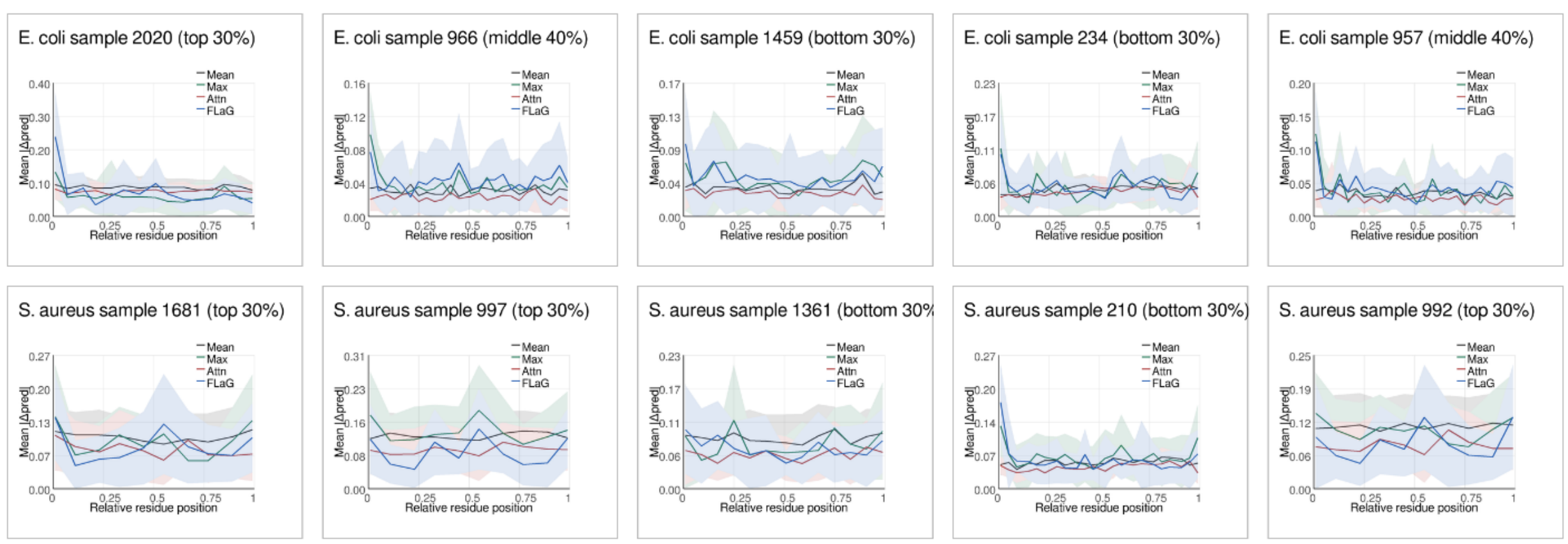

**Figure 4. Representative per-sample single-residue knockout response curves on the ESM2-8M AMP tasks, aggregated over 10 aligned seeds. The 10 peptides are arranged in the same 2 x 5 layout as in Figures 2, with *E. coli* on the top row and *S. aureus* on the bottom row. Within each panel, the horizontal axis shows the relative residue position along the peptide sequence and the vertical axis shows the mean absolute sequence-level prediction change after masking one residue position at a time in the final hidden layer. The colored curves correspond to mean pooling, max pooling, latent attention, and FLaG. The text above each panel identifies the bacterium-specific AMP task and the sample identifier; parenthesized bucket labels, when shown, indicate the same helix-propensity buckets used in Section 4.3.5, defined by ranking the 30-peptide manifest subset within each bacterium and partitioning it into the lowest 30%, middle 40%, and highest 30% of helix-propensity scores. This is a hidden-state knockout diagnostic rather than a raw amino-acid mutation experiment. Mean pooling remains comparatively flat, while max pooling, attention pooling, and FLaG preserve more differentiated residue-response profiles.**

### 4.3.4 Latent-Query Readout

To characterize the latent side of FLaG, we inspect the raw cross-attention weights within the multi-head attention operation in Eq. 4, after averaging across attention heads and then across the 10 aligned seeds. For the implementation details please see Supplementary Material S5.3. The results are shown in Figure 5.

The latent-query probe is best understood as showing weak but visible structure rather than a flat pattern. After averaging across attention heads and then across the 10 aligned seeds, the resulting heatmaps still exhibit only modest variation in absolute magnitude, but they are not constant across frequency bins. The dominant signal is sample specificity: different peptides exhibit clearly different cross-attention allocation patterns across frequency bins. Within a peptide, different latent queries also show mildly differentiated band preferences, although this query-wise differentiation is weaker and less stable than the peptide-level heterogeneity.

A more specific local pattern is that the raw cross-attention score often assigns slightly less mass to the very first frequency bin, while some nearby or mid-frequency bins receive comparatively higher weights. This behavior is not inconsistent with the earlier band-knockout and gate results. The raw cross-attention score is a normalized routing weight within the latent readout, whereas the knockout probe measures the end-to-end prediction effect of removing a band and the gate probe measures the subsequent spectral amplification. In practice, the lowest-frequency band still carries by far the largest spectral energy and remains strongly amplified after gating, so it can dominate final prediction sensitivity even if the latent readout does not place its largest relative score on the very first frequency bin. A plausible interpretation is that the latent queries mildly de-emphasize the most DC-like component (the near-zero-frequency component that captures the global or slowly varying part of the sequence) in order to avoid collapsing all readout mass onto the single strongest bin, while still preserving the overall low-frequency dependence of the full module.

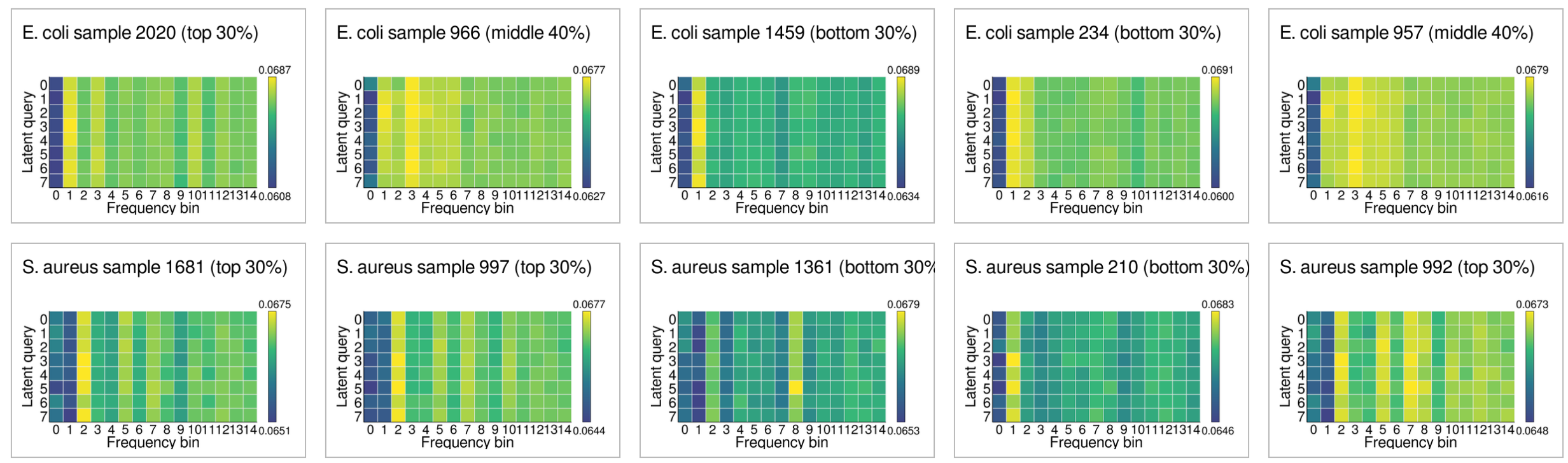

**Figure 5. Representative per-sample latent-query raw cross-attention-score heatmaps on the ESM2-8M AMP tasks, aggregated over 10 aligned seeds after first averaging across attention heads. The 10 peptides are arranged in the same 2 x 5 layout as in Figures 2, with *E. coli* on the top row and *S. aureus* on the bottom row. Within each panel, the horizontal axis shows the frequency bin and the vertical axis shows the latent-query index. Color indicates the averaged cross-attention score. The text above each panel identifies the bacterium-specific AMP task and the sample identifier; parenthesized bucket labels follow the same Section 4.3.5 helix-propensity stratification, obtained by ranking the 30-peptide manifest subset within each bacterium and partitioning it into the bottom 30%, middle 40%, and top 30% of helix-propensity values. The local views reveal weak but visible band-wise fluctuation together with clear sample-to-sample heterogeneity.**

## 4.3.5 Structure-Proxy Stratification

For this probe, we ask whether the peptide-level mechanism summaries co-vary with a lightweight structural proxy. For a peptide sequence $s_i$, we define the helix-propensity proxy as

$$h_i = \frac{1}{|s_i|} \sum_{a \in s_i} \mathbf{1}[a \in \{E, A, L, M, Q, K, R, H\}] \tag{19.}$$

Within each bacterium, the 30 sampled peptides are ranked by $h_i$ and split into bottom 30%, middle 40%, and top 30% buckets. Let $\boldsymbol{\mathcal{L}}$ denote the encoder layers and let $\boldsymbol{\mathcal{B}} = \{\boldsymbol{\mathcal{B}_1}, \boldsymbol{\mathcal{B}_2}, \dots, \boldsymbol{\mathcal{B}_8}\}$ denote the eight sequence-frequency bands, here each $\boldsymbol{\mathcal{B}}_b$ is the set of frequency bins assigned to band $b$. The per-peptide mean absolute band sensitivity shown in the first and third panels of Figure 6 is

$$S_i = \frac{1}{|\boldsymbol{\mathcal{L}}||\boldsymbol{\mathcal{B}}|} \sum_{\ell \in \mathcal{L}} \sum_{b \in \mathcal{B}} \left| \Delta MSE_{i,\ell,b} \right| \tag{20.}$$

Using the gate statistic from the previous subsection, the per-peptide gate-weight spread shown in the second and fourth panels of Figure 6 is

$$R_i = \max_b w_{i,b} - \min_b w_{i,b} \tag{21.}$$

Where $w_{i,b}$ is defined as

$$w_{i,b} = \frac{1}{|\mathcal{B}_b|} \sum_{f \in \mathcal{B}_b} \frac{e_{i,f}^{post}}{e_{i,f}^{pre} + \epsilon} \tag{22.}$$

Each bar in Figure 6 is the corresponding bucket-wise mean over peptides assigned to that helix-propensity bucket. The clearest stable trend is in band sensitivity: higher-helix buckets show stronger average sensitivity in both bacteria. By contrast, the gate trend is weaker and more dataset-dependent, so we treat this experiment as descriptive support for spectral sensitivity stratification rather than as evidence of biological mechanism.

Overall, the structure-proxy analysis is best interpreted as descriptive support for spectral sensitivity stratification, not as evidence of a resolved biological mechanism.

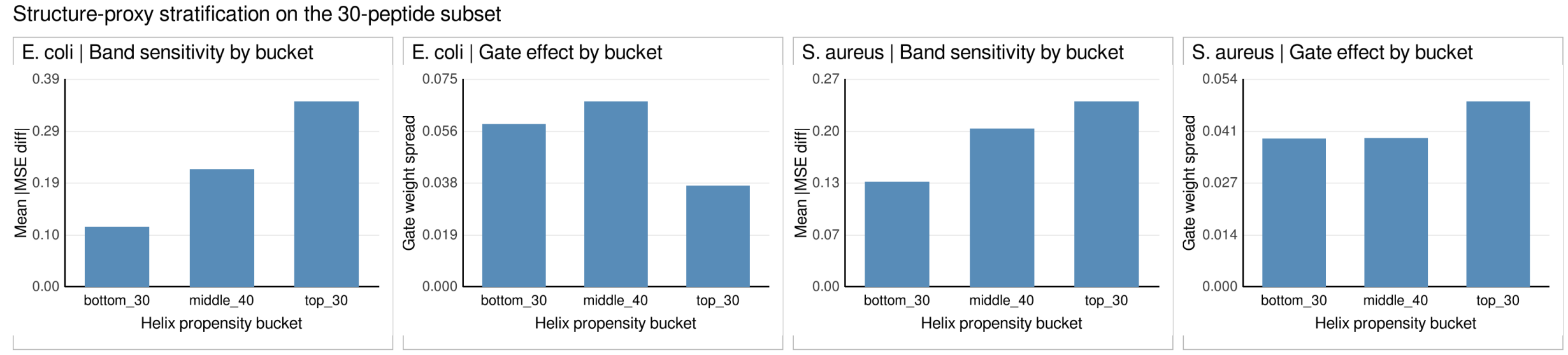

**Figure 6. Structure-proxy stratification on the 30-peptide subset for both AMP tasks. From left to right, the four panels show *E. coli* band sensitivity, *E. coli* gate-weight spread, *S. aureus* band sensitivity, and *S. aureus* gate-weight spread. Higher helix-propensity buckets exhibit stronger band sensitivity in both datasets, while the gate trend is milder and more dataset-dependent.**

Taken together, these probes point to a consistent antimicrobial peptide-side picture: low-frequency structure contributes the most overall, the residual contribution pattern of the remaining bands is more sample-specific, the gate broadly amplifies the spectrum while leaving the lowest-frequency bands most prominent, aggregation choice materially changes residue-level variation retention, latent readout patterns are sample-specific with mild query-wise differentiation, and stronger spectral sensitivity appears on more regular peptide subsets.

## 4.4 Ablation Study

**Table 6. Core cross-domain ablation over 10 aligned seeds. Antimicrobial peptide rows report RMSE (lower is better); image and text rows report accuracy (%). This aligned 10-seed comparison is the primary ablation used for model selection. Bold indicates the best or tied-best value in each row; underline indicates the second best.**

| Task | mean | latent | FFT+gate | FLaG |
|---|---|---|---|---|
| *E. coli* RMSE ↓ | 0.578 | 0.577 | **0.553** | <u>0.562</u> |
| *S. aureus* RMSE ↓ | 0.577 | 0.570 | <u>0.555</u> | **0.545** |
| CIFAR-100 Acc. ↑ | 76.78 | 75.71 | <u>77.09</u> | **77.20** |
| IMDB Acc. ↑ | **94.08** | 93.95 | <u>94.07</u> | **94.08** |

Table 6 is the primary ablation used for model selection because all four variants are evaluated under the same aligned 10-seed protocol across antimicrobial peptide, vision, and text. Under this controlled comparison, FLaG is best on *S. aureus* and CIFAR-100, slightly ahead on IMDB, and only behind FFT+gate on *E. coli*. We therefore interpret Table 6 as evidence about which variant serves as the default full-model choice under an aligned cross-domain comparison, rather than as a search for a simplified block that wins every individual row.

FFT+gate remains the closest reduced control because it shows that spectral gating accounts for a large fraction of the gain. Its advantage on *E. coli* indicates that channel-wise spectral modulation is a major contributor on that task. However, once model selection is restricted to the aligned 10-seed comparison in Table 6, the complete FLaG block remains the strongest overall choice because it leads on the harder antimicrobial peptide task and on CIFAR-100 while remaining competitive on text. Accordingly, we treat FFT+gate as the strongest reduced control and FLaG as the primary full variant within this aligned cross-domain comparison.

The remaining module-level ablations are reported in the supplementary material S1. Those lighter 5-seed sweeps are included to clarify component roles rather than to drive model selection, and they should not be interpreted as overturning the aligned 10-seed comparison in Table 6. Accordingly, the S1 tables are discussed as supporting diagnostics rather than as competing evidence against the primary ablation.

# 5. Limitations

FLaG has several limitations that define the current scope of the method. First, it assumes regularly sampled, padded token sequences: the current implementation applies the FFT to zero-masked batches, which is practical for variable-length inputs but may introduce padding-dependent spectral artifacts when true length variation is large, and irregularly sampled sequences would require alternatives such as the Non-uniform FFT [23]. Second, the FFT block adds computational overhead relative to lightweight token-domain pooling rules, so a fuller cross-domain efficiency study remains future work. Third, corruption-side behavior is perturbation-dependent rather than a general advantage, with mean pooling stronger under motion blur and contrast shift in the supplementary analysis.

# 6. Conclusion

We introduced FLaG, a frequency-domain sequence pooling method that transforms token representations via FFT, applies learnable latent attention to identify task-relevant frequency components, and modulates the spectrum through a gating mechanism before time-domain pooling. Evaluated across antimicrobial peptide, vision, and language tasks, FLaG achieves its clearest gains on the 8M antimicrobial peptide setting and on CIFAR-100, while remaining competitive with mean pooling and RoBERTa-native sentence representations on clean IMDB, and more mixed on GLUE. The auxiliary probes and ablations refine the central claim: on the antimicrobial peptide side, FLaG is most sensitive to coarse sequence-scale structure, uses a broadly shared spectral reweighting stage rather than a sharply selective gate, and preserves more differentiated residue-response profiles than mean pooling, Taken together, these results suggest that the complete FLaG block is best viewed as a competitive full-model choice whose strongest evidence currently comes from structure-sensitive antimicrobial peptide settings and CIFAR-100.

# Acknowledgements

This study was supported by the Fundamental Research Funds for the Central Universities (JLU).

# References

[1] C. Lee *et al.*, “NV-embed: Improved techniques for training LLMs as generalist embedding models,” arXiv.org. Accessed: May 26, 2026. [Online]. Available: https://arxiv.org/abs/2405.17428v3

[2] Z. Lin *et al.*, “A structured self-attentive sentence embedding,” in *Proc. 5th int. Conf. Learn. Represent. (ICLR)*, 2017. [Online]. Available: https://openreview.net/forum?id=BJC_jUqxe

[3] J. Lee, Y. Lee, J. Kim, A. Kosiorek, S. Choi, and Y. W. Teh, “Set transformer: A framework for attention-based permutation-invariant neural networks,” in *Proceedings of the 36th International Conference on Machine Learning*, PMLR, May 2019, pp. 3744–3753. Accessed: Mar. 17, 2026. [Online]. Available: https://proceedings.mlr.press/v97/lee19d.html

[4] Y. Tang and Y. Yang, “Pooling and attention: What are effective designs for LLM-based embedding models?,” *arXiv preprint arXiv:2409.02727*, 2024, doi: 10.48550/arXiv.2409.02727.

[5] N. NaderiAlizadeh and R. Singh, “EvoPool: Evolution-guided pooling of protein language model embeddings,” *openRxiv*, 2026, doi: 10.64898/2026.02.02.703349.

[6] N. Bonneel, J. Rabin, G. Peyré, and H. Pfister, “Sliced and radon wasserstein barycenters of measures,” *J. Math. Imaging Vis.*, vol. 51, no. 1, pp. 22–45, 2015, doi: 10.1007/s10851-014-0506-3.

[7] N. Rahaman *et al.*, “On the spectral bias of neural networks”.

[8] T. N. Kipf and M. Welling, “Semi-supervised classification with graph convolutional networks,” in *Proc. 5th int. Conf. Learn. Represent. (ICLR)*, 2017. [Online]. Available: https://openreview.net/forum?id=SJU4ayYgl

[9] T. Zhou, Z. Ma, Q. Wen, X. Wang, L. Sun, and R. Jin, “FEDformer: Frequency enhanced decomposed transformer for long-term series forecasting,” in *Proc. 39th int. Conf. Mach. Learn. (ICML)*, in Proc. Mach. Learn. Res., vol. 162. 2022, pp. 27268–27286. [Online]. Available: https://proceedings.mlr.press/v162/zhou22g.html

[10] Z. Lin *et al.*, “Evolutionary-scale prediction of atomic-level protein structure with a language model,” *Science*, vol. 379, no. 6637, pp. 1123–1130, 2023, doi: 10.1126/science.ade2574.

[11] Z. Xu, J. Wu, Y. S. Song, and R. Mahadevan, “Enzyme activity prediction of sequence variants on novel substrates using improved substrate encodings and convolutional pooling,” in *Proceedings of the 16th Machine Learning in Computational Biology meeting*, PMLR, Jan. 2022, pp. 78–87. Accessed: May 26, 2026. [Online]. Available: https://proceedings.mlr.press/v165/xu22a.html

[12] A. Tartici, G. Nayar, and R. B. Altman, “Pool PaRTI: A PageRank-based pooling method for identifying critical residues and enhancing protein sequence representations,” *bioRxiv*, p. 2024.10.04.616701, Mar. 2025, doi: 10.1101/2024.10.04.616701.

[13] Y. Liu *et al.*, “RoBERTa: A robustly optimized BERT pretraining approach,” *arXiv preprint arXiv:1907.11692*, 2019, doi: 10.48550/arXiv.1907.11692.

[14] “Foundations of computer vision.” Accessed: Jun. 02, 2026. [Online]. Available: https://mitpress.mit.edu/9780262048972/foundations-of-computer-vision/

[15] A. Jaegle *et al.*, “Perceiver IO: A general architecture for structured inputs and outputs,” in *Proc. 10th int. Conf. Learn. Represent. (ICLR)*, 2022. [Online]. Available: https://openreview.net/forum?id=fILj7WpI-g

[16] K. Li *et al.*, “AMPCliff: Quantitative definition and benchmarking of activity cliffs in antimicrobial peptides,” *Journal of Advanced Research*, 2025.

[17] J. Hyun, H. Seong, and E. Kim, “Universal pooling – a new pooling method for convolutional neural networks,” *Expert Systems with Applications*, vol. 180, p. 115084, Oct. 2021, doi: 10.1016/j.eswa.2021.115084.

[18] K. He, X. Zhang, S. Ren, and J. Sun, “Deep residual learning for image recognition,” in *Proc. IEEE conf. Comput. Vis. Pattern recognit. (CVPR)*, 2016, pp. 770–778. doi: 10.1109/CVPR.2016.90.

[19] A. Krizhevsky, “Learning multiple layers of features from tiny images,” Univ. of Toronto, 2009. [Online]. Available: https://www.cs.toronto.edu/~kriz/learning-features-2009-TR.pdf

[20] A. L. Maas, R. E. Daly, P. T. Pham, D. Huang, A. Y. Ng, and C. Potts, “Learning word vectors for sentiment analysis,” in *Proc. 49th annu. Meeting assoc. Comput. Linguist. (ACL)*, 2011, pp. 142–150. [Online]. Available: https://aclanthology.org/P11-1015/

[21] A. Wang, A. Singh, J. Michael, F. Hill, O. Levy, and S. R. Bowman, “GLUE: A multi-task benchmark and analysis platform for natural language understanding,” in *Proc. 2018 EMNLP workshop BlackboxNLP: Analyzing and interpreting neural networks for NLP*, 2018, pp. 353–355. doi: 10.18653/v1/W18-5446.

[22] A. Tamkin, D. Jurafsky, and N. Goodman, “Language through a prism: A spectral approach for multiscale language representations,” in *Advances in Neural Information Processing Systems*, Curran Associates, Inc., 2020, pp. 5492–5504. Accessed: Oct. 24, 2025. [Online]. Available: https://proceedings.neurips.cc/paper_files/paper/2020/hash/3acb2a202ae4bea8840224e6fce16fd0-Abstract.html

[23] L. Greengard and J.-Y. Lee, “Accelerating the nonuniform fast fourier transform,” *SIAM Rev.*, vol. 46, no. 3, pp. 443–454, 2004, doi: 10.1137/S003614450343200X.